# A French Corpus Annotated for Multiword Expressions with Adverbial Function

**Eric Laporte, Takuya Nakamura, Stavroula Voyatzi**
Université Paris-Est
Institut Gaspard-Monge - LabInfo
5, Boulevard Descartes, Champs-sur-Marne
77454 Marne-la-Vallée Cedex 2 (France)
E-mail: eric.laporte@univ-paris-est.fr, nakamura@univ-mlv.fr, voyatzi@univ-mlv.fr

**Abstract**

This paper presents a French corpus annotated for multiword expressions (MWEs) with adverbial function. This corpus is designed for investigation on information retrieval and extraction, as well as on deep and shallow syntactic parsing. We delimit which kind of MWEs we annotated, we describe the resources and methods we used for the annotation, and we briefly comment the results. The annotated corpus is available at *http://infolingu.univ-mlv.fr/* under the LGPLLR license.

## 1. Introduction

Recognising multiword adverbs such as *à long terme* 'in the long run' in texts is likely to be useful for information retrieval and extraction because of the information that such adverbials can convey. In addition, it is likely to help resolving prepositional attachment during shallow or deep parsing: most multiword adverbs have the superficial syntax of prepositional phrases; in many cases, recognising them rules out analyses where they are arguments or noun modifiers.

The quality of the recognition of multiword adverbs depends on algorithms, but also on resources. We created a corpus of French texts annotated with multiword adverbs. In this article, we survey related work, we define the target of our annotation effort, we describe the method we have implemented and we analyse the corpus obtained. This corpus will be made freely available on the web under the LGPLLR license when this article is published.

## 2. Related work

Corpora annotated with multiword adverbs are rare and small[1]. In the Grace corpus (Rajman *et al.*, 1997), most multiword units are ignored. In the French Treebank (Abeillé *et al.*, 2003), prepositional phrases and adverbs are annotated with a binary feature ('compound') which indicates whether they are multiword units; the distinction between whether prepositional phrases are verb modifiers, noun modifiers or objects appears only in the function-annotated part of the Treebank (350 000 words). We are not aware of other available French corpora annotated with multiword adverbs. In other languages, including English, corpora annotated with multiword units are rare and small as well.

## 3. Target of annotation

The target of our annotation effort is defined by the intersection of two criteria: (i) multiword expressions and (ii) adverbial function. In this section, we define both criteria in more detail, we define the features that we included in the annotations, and we describe the corpus.

### 3.1 Multiword expression criterion

For this work, we considered a phrase composed of several words to be a multiword expression if some or all of their elements are frozen together in the sense of (Gross, 1986), that is, if their combination does not obey productive rules of syntactic and semantic compositionality. In the following example, *de nos jours* ('nowadays', lit. 'of our days') is a multiword adverb:

(1) *Il est facile* ***de nos jours*** *de s'informer*
'It is easy to get informed **nowadays**'

This criterion ensures a complementarity between lexicon and grammar. In other words, it tends to ensure[2] that any combination of linguistic elements which is licit in the language, but is not represented in syntactic-semantic grammars, will be stored in lexicons.

Syntactic-semantic compositionality is usually defined as follows (Freckleton, 1985; Machonis, 1985; Silberztein, 1993; Lamiroy, 2003): a combination of linguistic elements is compositional if and only if its meaning can be computed from its elements. This is also our conception. However, in this definition, we consider that the possibility of computing the meaning of phrases from their elements is of any interest only if it is a better solution than storing the same phrases in lexicons, i.e. if they rely on grammatical rules with sufficient generality. In other words, we consider a combination of linguistic elements to be compositional if and only if its meaning can be computed from its elements **by a grammar**. In example (1) above, the lack of compositionality is apparent from distributional restrictions[3] such as:

> * *Il est facile de nos **semaines** de s'informer*
> *'It is easy to get informed nowa**weeks**'

[1] Several reasons explain this lack of interest. Firstly, adverbials are usually felt as less useful than nouns for information retrieval and extraction. Secondly, many multiword adverbs are difficult to distinguish from prepositional phrases assuming other syntactic functions, such as arguments or noun modifiers: the distinction is hardly correlated to any material markers in texts and lies in complex linguistic notions (Villavicencio, 2002; Merlo, 2003). The task is therefore felt as too difficult by most researchers in language processing, whose main background is in information technology. However, the distinction in question is essential to identifying the semantic core of a sentence, and the availability of a larger corpus of annotated text is likely to shed light on the problems posed by this task.

[2] That can be empirically checked only after a lexicon and a grammar for the same language are complete and compatible.

Multiword expressions include many different subtypes, varying from entirely fixed expressions to syntactically more flexible expressions (Sag *et al*., 2002). We annotated expressions undergoing variations[4]. In (2), the possessive adjective agrees obligatorily in person and number with the subject of the sentence:

(2) ***De (ses + *mes) propres mains**, il a construit une maison*
'**With (his + *my) own hands**, he built a house'

## 3.2 Adverbial function

We annotated only expressions with adverbial function, or circumstantial complements, i.e. complements which are not objects of the predicate of the clause in which they appear. We recognised them through criteria (Gross 1986, 1990a, 1990b) involving the fact that they are optional, they combine freely with a wide variety of predicates and some of them pronominalize with specific forms. Phrases with adverbial function are often called 'circumstantial complements', 'adverbials', 'adjuncts', or 'generalised adverbs'. They assume several morphosyntactic forms: underived (*demain* 'tomorrow') or derived adverbs (*prochainement* 'soon'), prepositional phrases (*à la dernière minute* 'at the last minute') or circumstantial clauses (*jusqu'à ce que mort s'ensuive* 'until death comes'), and special structures in the case of named entities of time (*lundi 20* 'on Monday 20'). We annotated NEs only when they have an adverbial function, as in: *Jean arrive lundi 20* 'John arrives on Monday 20'. NEs of other categories, such as places, persons, events, etc., are usually not adverbials.

## 3.3 Features

Two types of features were included in the annotations.
(i) Each occurrence of a multiword adverb was assigned one internal morphosyntactic structure or semantic type among 19. The definition of the morphosyntactic structures is based on the number, category and position of the frozen and free components of the adverbial. They are described as a sequence of parts of speech and syntactic categories. For example, *à la nuit tombante* 'at nightfall' is assigned a structure identified by the mnemonic acronym *PCA*, and defined as *Prép Dét C (MPA) Adj*, where *C* stands for a noun frozen with the rest of the adverbial, *Adj* for a post-posed noun modifier (e.g. an adjectival phrase or a relative clause), and *MPA* for a pre-adjectival modifier, empty in this lexical item. For named entities, this feature encodes the semantic type: date, duration, time or frequency, in conformity with the typology of the Infom@gic project (Martineau *et al.*, 2007). The 19 structures and semantic types are listed in Table 1. In this table, *N* stands for a free noun phrase, and *W* for a variable ranging over verb complements. Other symbols are easy to interpret: *Prép*, *Dét*, *Adj*, *V*, *Conj*...

| Identifiers | Structures | Examples |
|---|---|---|
| PC | Prép C | *en bref* |
| PDETC | Prép Dét C | *de nos jours* |
| PAC | Prép Adj C | *à la dernière minute* |
| PCA | Prép C Adj | *à la nuit tombante* |
| PCDC | Prép C de C | *dans la limite du possible* |
| PCPC | Prép C Prép C | *des pieds à la tête* |
| PCONJ | Prép C Conj C | *en tout et pour tout* |
| PCDN | Prép C de N | *au moyen de N* |
| PCPN | Prép C Prép N | *par rapport à N* |
| PV | Prép V W | *à dire vrai* |
| PF | P (frozen clause) | *jusqu'à ce que mort s'ensuive* |
| PECO | (Adj) comme C | *comme ses pieds* |
| PVCO | (V) comme C | *comme un cheveu sur la soupe* |
| PPCO | (V) comme Prép C | *comme dans du beurre* |
| PJC | Conj C | *mais enfin et surtout* |
| DATE | Named Entities | *le 22 mai 2008* |
| DURATION | Named Entities | *pendant vingt-quatre heures* |
| TIME | Named Entities | *à huit heures du soir* |
| FREQUENCE | Named Entities | *deux fois par jour* |

Table 1: Morphosyntactic structures and semantic types of MWEs with adverbial function

(ii) The second feature is binary and encodes whether the adverbial assumes a conjunctive function in discourse, i.e. it connects the clause in which the adverbial occurs with the previous clause, as *en dernier lieu* 'finally'. The positive value is indicated by identifier 'Conj' in attribute 'fs'. Example: *<ADV fs='PAC Conj'>*.

## 3.4 The corpus

The corpus we annotated includes: (a) the complete minutes of the sessions of the French National Assembly on October 3-4, 2006, transcribed into written style from oral French (hereafter *AS*)[5] and (b) Jules Verne's novel *Le Tour du monde en quatre-vingts jours*, 1873 (hereafter *JV*). Errors (e.g. *mis enoeuvre* for *mis en oeuvre* 'implemented') have not been corrected. Statistics on the corpus are displayed in Table 2.

---

[3] The point is that this blocking of distributional variation (and other syntactic constraints) cannot be predicted on the basis of general grammar rules and independently needed lexical entries. Therefore, the acceptable combinations are meaning units and have to be included in lexicons as multiword lexical items.

[4] We annotated phrases which comprise a frozen part and a free part, e.g. *au moyen de ce bouton* 'with the aid of this switch', in which *au moyen de* 'with the aid of' is frozen, and *ce bouton* 'this switch' is a distributionally free noun phrase embedded in the global phrase. In such cases, we delimited the embedded free part with tags (cf. section 4.2). Finally, we annotated named entities (NEs) of date and duration. The status of named entities with respect to compositionality is not fully consensual: however, we complied with the usual view that, since they follow quite specific grammatical rules, they should be considered as multiword expressions.

[5] http://www.assemblee-nationale.fr/12/documents/index-rapports.asp.

| | size (Kb) | sentences | tokens | types |
|---|---|---|---|---|
| **corpus *AS*** | 824 | 5 146 | 98 969 | 18 028 |
| **corpus *JV*** | 1 231 | 3 648 | 69 877 | 19 828 |
| **total** | 2 055 | 8 794 | 168 846 | 37 856 |

Table 2: Size of the corpus

# 4. Methodology

In order to annotate the corpus, we tagged the occurrences of the expressions described in a syntactic-semantic lexicon of adverbials, as Abeillé *et al.* (2003), Baptista (2003) for Portuguese, and Català & Baptista (2007) for Spanish; we tagged NEs of date, duration, time, and frequency through a set of local grammars, as Friburger & Maurel (2004); then, we revised the annotation manually.

## 4.1 The lexicon

We used the same syntactic-semantic lexicon (Gross, 1990a) as Abeillé *et al.* (2003), so that the two corpora can be used jointly for further research. This lexicon has 6 800 entries. It is freely available[6] for research and business under the LGPLLR license. It was constructed on the basis of conventional dictionaries, grammars, corpora and introspection, within the Lexicon-Grammar methodology (Gross, 1986; 1994). It takes the form of a set of Lexicon-Grammar tables such that of Table 3, which displays a sample of the lexical items with the *PCA* morphosyntactic structure.

| | A | B | C | D | E | F | G | H | I | J | K | L | M | N |
|---|---|---|---|---|---|---|---|---|---|---|---|---|---|---|
| 1 | N0 =: Nhum | N0 =: N-hum | Nég obl | | | Prép | Dét | C | Modif pré-adj | Adj | Prép Dét C | Prép Dét MPA Adj C | Prép MPA Adj C | Conjonction |
| 170 | + | - | - | <E> | agir | dans | les | délais | les plus | brefs | - | - | + | - |
| 171 | + | - | - | <E> | agir | dans | les | délais | les plus | courts | - | - | + | - |
| 172 | + | - | - | <E> | agir | dans | les | délais | les | meilleurs | - | - | + | - |
| 173 | + | - | - | <E> | rire | <E> | toutes | dents | <E> | dehors | - | - | - | - |
| 174 | - | + | - | :se | produire | à | cette | époque- | <E> | ci | + | - | - | - |
| 175 | - | + | - | :se | produire | à | cette | époque- | <E> | là | + | - | - | + |

Table 3: Sample of the table of entries with the *PCA* morphosyntactic structure

In this table, each row describes a lexical item, and each column corresponds:

- either to one of the elements in the morphosyntactic structure of the items (columns with identifiers 'Prép', 'Dét', 'C', 'Modif pré-adj' and 'Adj');

- or to a syntactic-semantic feature (columns with binary values), for example the conjunctive function of the adverbial in discourse (column with identifier 'Conjonction'), or the constraint that the adverbial obligatorily occurs in a negative clause (column with identifier 'Nég obl');

- or to illustrative information provided as an aid for the human reader to find examples of sentences containing the adverbial (e.g. columns D and E giving an example of a verb compatible with the adverb).

There are 15 such tables, one for each of the morphosyntactic structures. The features provided by the lexicon were used to annotate the occurrences.

[6] http://infolingu.univ-mlv.fr/english/DonneesLinguistiques/Lexiques-Grammaires/View.html.

## 4.2 Tagging

We tagged the corpus with the Unitex system (Paumier, 2006). Many multiword adverbs are entirely fixed expressions, but others present variations, such as grammatical agreement (cf. example (2), section 3.1), permutations and omissions. Due to these variations, we tagged them with finite-state transducers (FST): the input part of these transducers recognises the expressions and their variants, and the output part inserts the tags. Like Català & Baptista (2007), we used lexicalised transducers, i.e. one for each lexical item, and we generated them with the technique of parameterised graphs (Roche, 1999) modified by Silberztein (1999).

Multiword adverbs with a free prepositional phrase modifier (morphosyntactic structures *PCDN* and *PCPN*) were annotated semi-automatically as follows ('*N*' if the free complement is occupied by a noun phrase, '*S*' if it is occupied by a clause):

(i) *<ADV fs='PCDN'>**compte tenu de** <NP>**vos ambitions**</NP></ADV>*
'taking into account your ambitions'

(ii) *<ADV fs='PCDN'>**compte tenu de** <S>**ce que tout va bien**</S></ADV>*
'taking into account that everything is OK'

Named entities with temporal value (cf. section 3.2) were automatically tagged by using FST methods similar to those applied for multiword adverbs.

## 4.4 Manual revision

The annotation was manually reviewed by three experts.

This validation followed guidelines, which are available along with the corpus. It involved two operations.

(i) The sequences tagged with the aid of the lexicon and Unitex were checked in order to detect cases when the recognised sequence is in fact a part of a larger MWE. For instance, when *de force* 'forcibly' occurred within the compound noun *ligne de force* 'thrust', the tags around *de force* were deleted.

When the embedded free part of a multiword adverb is a coordination, we tagged it manually:

*<ADV fs='PCDN'>en termes de <NP>santé</NP> et d'<NP>éducation</NP></ADV>*
'in terms of health and education'

(ii) The text was integrally reviewed in search for multiword adverbs absent from the lexicon, and thus undetected by Unitex, e.g. *de plus* 'moreover' or *pour le moins* 'at least'.

This required for the annotators to identify the syntactic structure of each sentence in the corpus. We had meetings during the annotation process in order to make it consistent.

## 5. Results

This corpus is annotated with 4 247 occurrences of MWEs with adverbial function. They represent about 6 % of the overall of simple word occurrences occurring in the whole corpus. Table 4, below, shows the number of occurrences of annotated MWEs. The lines of the table correspond to the morphosyntactic structures and semantic types.

| identifiers | *JV* corpus | % *JV* cover | *AS* corpus | % *AS* cover |
|---|---|---|---|---|
| PC | 338 | 1.38 | 420 | 1.28 |
| PDETC | 257 | 1.16 | 165 | 0.64 |
| PAC | 77 | 0.35 | 127 | 0.51 |
| PCA | 55 | 0.30 | 53 | 0.22 |
| PCDC | 38 | 0.17 | 36 | 0.12 |
| PCPC | 37 | 0.15 | 59 | 0.20 |
| PCONJ | 13 | 0.07 | 21 | 0.08 |
| PCDN | 248 | 1.00 | 834 | 2.52 |
| PCPN | 103 | 0.41 | 107 | 0.32 |
| PV | 53 | 0.21 | 54 | 0.17 |
| PF | 11 | 0.04 | 23 | 0.07 |
| PECO | 1 | 0.00 | 1 | 0.00 |
| PVCO | 8 | 0.04 | 3 | 0.00 |
| PPCO | 2 | 0.00 | 1 | 0.00 |
| PJC | 2 | 0.00 | 3 | 0.00 |
| DATE | 258 | 1.00 | 383 | 1.04 |
| DURATION | 120 | 0.49 | 111 | 0.31 |
| TIME | 128 | 0.50 | 29 | 0.06 |
| FREQUENCE | 31 | 0.11 | 37 | 0.10 |
| Total | 1 780 | 6.28 | 2 467 | 6.64 |

Table 4: Annotated occurrences of MWEs with adverbial function in the corpus

## 6. Conclusion

This paper described the design of a French corpus annotated for MWEs with adverbial function. Various types of features are included in the annotations: the morphosyntactic structure, special functions in discourse (e.g. the conjunctive function) and the semantic types of named entities of time. This annotated corpus can be used jointly with the French Treebank (Abeillé *et al*., 2003) for research on information retrieval and extraction, automatic lexical acquisition, as well as on deep and shallow syntactic parsing.

## 7. Acknowledgment

This task has been partially financed by CNRS and by the Cap Digital business cluster. We thank Anne Abeillé for making the French Treebank available to us.

## 8. References

Abeillé, A., Clément, L., and Toussenel F. (2003). Building a Treebank for French. In A. Abeillé (Ed.), *Building and Using Parsed Corpora*, *Text, Speech and Language Technology*, 20, Kluwer, Dordrecht, pp. 165--187.

Baptista, J. (2003). Some Families of Compound Temporal Adverbs in Portuguese. In *Proceedings of the Workshop on Finite-State Methods for Natural Language Processing, EACL 2003*, Budapest, Hungary, pp. 97--104.

Català, D., Baptista, J. (2007). Spanish Adverbial Frozen Expressions. In *Proceedings of the Workshop on a Broader Perspective on Multiword Expressions, ACL 2007*, Prague, Czech Republic, pp. 33--40.

Freckleton, P. (1985). Sentence idioms in English, *Working Papers in Linguistics*, University of Melbourne, pp. 153--168 & appendix (196 p.).

Friburger, N., Maurel, D. (2004). Finite-state transducer cascades to extract named entities in texts. *Theoretical Computer Science*, 313(1), pp. 93--104.

Gross, M. (1986). Lexicon-Grammar. The representation of compound words. In *Proceedings of the Eleventh International Conference on Computational Linguistics*, Bonn, West Germany, pp. 1--6.

Gross, M. (1990a). *Grammaire transformationnelle du français: 3. Syntaxe de l'adverbe*. Paris, ASSTRIL.

Gross, M. (1990b). La caractérisation des adverbes dans un lexique-grammaire. *Langue Française*, 86, pp. 90--102.

Gross, M. (1994). Constructing Lexicon-Grammars. In Atkins & Zampoli (Eds.), *Computational Approaches to the Lexicon*, Oxford University Press, pp. 213--263.

Lamiroy, B. (2003). Les notions linguistiques de figement et de contrainte, *Lingvisticae Investigationes*, 26:1, Amsterdam/Philadelphia: John Benjamins, pp. 1--14.

Machonis, P. (1985). Transformations of verb phrase idioms: passivization, particle movement, dative shift. *American Speech*, 60:4, pp. 291--308.

Martineau, C., Tolone, E., Voyatzi, S. (2007). Les Entités Nommées : usage et degrés de précision et de désambiguïsation. In *Proceedings of the Twenty Sixth International Conference on Lexis and Grammar*, Bonifacio, Corse du Sud, pp. 105--112.

Merlo, P. (2003). Generalised PP-attachment Disambiguation using Corpus-based Linguistic Diagnostics. In *Proceedings of the Tenth Conference of the European Chapter of the Association for Computational Linguistics*, Budapest, Hungary, pp. 251--258.

Paumier, S. (2006). *Unitex Manual*. Université Paris-Est. http://igm.univ-mlv.fr/~unitex/manuel.html.

Rajman, M., Lecomte, J., Paroubek, P. (1997). Format de description lexicale pour le français. Partie 2 : Description morpho-syntaxique. Rapport GRACE GTR-3--2.1.

Roche, E. (1999). Finite-state transducers: parsing free and frozen sentences. In Kornai (Ed.), *Extended finite-state models of language*, Cambridge University Press, pp. 108--120.

Sag, I. A., Baldwin, T., Bond, F., Copestake, A., Flickinger, D. (2002). Multiword Expressions: A Pain in the Neck for NLP. In A. Gelbuk (Ed.), *Computational Linguistics and Intelligent Text Processing: Proceedings of the Third International Conference CICLing 2002*, Springer-Verlag, Heidelberg/Berlin, pp. 1--15.

Silberztein, M. D. (1993) Les groupes nominaux productifs et les noms composés lexicalisés. *Lingvisticae Investigationes*, 17:2, Amsterdam/ Philadelphia, John Benjamins, pp.405--426.

Silberztein, M. (1999). *Manuel d'utilisation d'Intex version 4.12*.

Villavicencio, A. (2002). Learning to distinguish PP arguments from adjuncts. In *Proceedings of the Sixth Conference on Natural Language Learning*, Taipei, Taiwan, pp. 84--90.